\documentclass[journal]{IEEEtran}

\usepackage[ngerman,english]{babel}

\usepackage{ucs}

\usepackage[utf8]{inputenc}

\usepackage{comment}

\usepackage{enumerate}

\usepackage[babel]{csquotes}

\usepackage{cite}

\usepackage{fixmath}

\usepackage{lmodern}

\usepackage{amsmath}
\usepackage{amssymb}

\usepackage{graphicx}

\usepackage{epstopdf}

\usepackage{tikz}

\usepackage{pgfplots}

\usepackage{hhline}

\usepackage{hyperref}

\usepackage{makecell}

\usepackage{array}

\usepackage{microtype}

\DeclareMathOperator{\ep}{E}		

\DeclareMathOperator{\TransposedOp}{T}	

\newcommand{\tp}{{\TransposedOp}}

\DeclareMathAlphabet{\mathbit}{OML}{cmr}{bx}{it}
\DeclareMathAlphabet{\mathss}{T1}{cmss}{m}{sl}
\DeclareMathAlphabet{\mathssbold}{T1}{cmss}{bx}{sl}

\newcommand{\mb}[1]{\mathbit{#1}}  
\newcommand{\zero}{\boldsymbol{0}}
\newcommand{\id}{\mathbf{I}} 

\usepgfplotslibrary{fillbetween}

\newcommand{\thickhline}{%
	\noalign {\ifnum 0=`}\fi \hrule height 1pt
	\futurelet \reserved@a \@xhline
}
\newcolumntype{?}{@{\hskip\tabcolsep\vrule width 1pt\hskip\tabcolsep}}
\makeatother

\begin{document}

\title{A Comparison of Prediction Algorithms and Nexting for Short Term Weather Forecasts}
\author{Michael Koller, Johannes Feldmaier, Klaus Diepold\\ Dept. ECE, Technical University of Munich (TUM)}
\maketitle

\begin{abstract}
This report first provides a brief overview of a number of supervised learning algorithms for regression tasks.
Among those are neural networks, regression trees, and the recently introduced Nexting.
Nexting has been presented in the context of reinforcement learning where it was used to predict a large number of signals at different timescales.
In the second half of this report, we apply the algorithms to historical weather data in order to evaluate their suitability to forecast a local weather trend.
Our experiments did not identify one clearly preferable method but rather show that choosing an appropriate algorithm depends on the available side information.
For slowly varying signals and a proficient number of training samples, Nexting achieved good results in the studied cases.
\end{abstract}

\section{Introduction}
Typically, regression algorithms follow a two stage process.
In a first step, the algorithms are trained on historic data and optimized.
Then, they are used to predict outcomes based on the learned rules.
This approach is known as supervised learning.

In this report, we compare a number of such algorithms.
First, we briefly introduce them in Section~\ref{sec: algorithms} and illustrate their main idea by applying them to a sinusoidal signal.
One of the chosen algorithms is the recently introduced Nexting \cite{MoWhSu11}.
It was designed to predict numerous target signals at multiple timescales.
We compare Nexting to well-known methods like polynomial regression.

In Section~\ref{sec: application}, we apply the most promising candidates to historical weather measurements.
The results are compared to assess the suitability of the algorithms as a predictor of future data.
Being able to predict a local trend might be interesting for applications that depend on weather conditions.
For example, this could help to improve forecasts of the energy production of solar systems or wind power plants.
The last section then concludes our findings.

\section{Algorithms}
\label{sec: algorithms}
We write vectors like $ \mb{x} \in \mathbb{R}^P $ in bold face and denote their elements $ x_i $ by subscripts.

Consider a physical system whose output $ y \in \mathbb{R} $ can be described by a function $ f( \mb{x} ) = y $ where $ \mb{x} \in \mathbb{R}^P $ collects the $ P $ system inputs.
Assuming that the function $ f( \mb{x} ) $ is unknown, we try to approximate it by some $ \hat{f}( \mb{x} ) $.
To determine this approximation, we first take $ D+F $ measurements $ ( \mb{\xi}^{(i)}, y^{(i)} ) \in \mathbb{R}^P \times \mathbb{R} $ with $ y^{(i)} = f( \mb{\xi}^{(i)} ) $, $ i = 1, 2, \dots, D, D+1, \dots, D+F $. 
Then, we apply the regression methods described below to the data set $ \{ ( \mb{\xi}^{(i)}, y^{(i)} ) \}_{i=1}^D $, which yields $ \hat{f}( \mb{x} ) $.
Having done so, we are interested in the behavior of $ \hat{f}(\mb{x}) $ for input $ \mb{\xi}^{(i)} $ with $ i > D $.
We thus aim at assessing the suitability of $ \hat{f}(\mb{x}) $ as a predictor of future data $ y^{(D+1)}, y^{(D+2)}, \dots, y^{(D+F)} $.

In the special case that $ P = 1 $ and $ f( x )  = f( t ) $ is a function of time $ t $ which we sample with period $ T $, we write $ y[ t+j ] = f( t+jT ) $ for integers $ j $.
This is a convenient notation for some of the following algorithms and in case of periodic signals.

\subsection{Polynomial Regression}
In polynomial regression, we model the unknown function $ f(x) $ by a polynomial
	\begin{equation}
	f_{\mathrm{PR}} (x) = \sum_{l=0}^{L} \theta_l x^l = \theta_0 + \theta_1 x + \theta_2 x^2 + \dots + \theta_L x^L
	\end{equation}
of degree $ L $.
Note that $ x \in \mathbb{R} $ in this case.
Given our training data $ \{ (\xi^{(i)}, y^{(i)}) \}_{i=1}^D $, we have the system of equations
	\begin{align}
	\label{eq: pol reg system of eq}
	\underbrace{
	\begin{bmatrix}
		y^{(1)} \\ y^{(2)} \\ \vdots \\ y^{(D)}
	\end{bmatrix}}_{\mb{y}} =
	\underbrace{
	\begin{bmatrix}
		1 & \xi^{(1)} & \xi^{(1),2} & \dots & \xi^{(1),L} \\
		1 & \xi^{(2)} & \xi^{(2),2} & \dots & \xi^{(2),L} \\
		\vdots & \vdots & \vdots & & \vdots \\
		1 & \xi^{(D)} & \xi^{(D),2} & \dots & \xi^{(D),L}
	\end{bmatrix}}_{\mb{X}}
	\underbrace{
	\begin{bmatrix}
		\theta_0 \\ \theta_1 \\ \vdots \\ \theta_L
	\end{bmatrix}}_{\mb{\theta}}.
	\end{align}
To fit the polynomial coefficients $ \mb{\theta} $, we use the least squares approach, i.e., we solve the optimization problem
	\begin{equation}
	\hat{\mb{\theta}}_{\mathrm{PR}} = \arg \min_{\mb{\theta}} \| \mb{y} - \mb{X} \mb{\theta} \|_2^2
	\end{equation}
where $ \| . \|_2 $ denotes the Euclidean norm.
Writing the norm as an inner product, setting the derivative with respect to $ \mb{\theta} $ to zero, and solving for $ \mb{\theta} $ leads to the unique solution
	\begin{equation}
	\label{eq: pol reg solution}
	\hat{\mb{\theta}}_{\mathrm{PR}} = ( \mb{X}^\tp \mb{X} )^{-1} \mb{X}^\tp \mb{y}
	\end{equation}
provided that the inverse $ ( \mb{X}^\tp \mb{X} )^{-1} $ exists (cf.~\cite{HaTiFr09}).

Summing up the above, our approximating function is
	\begin{equation}
	\label{eq: pol reg function}
	\hat{f}_{\mathrm{PR}} (x) = \sum_{l=0}^{L} \hat{\theta}_{l,\mathrm{PR}} x^l.
	\end{equation}
Predicting the $ y^{(i)} $ is therefore done by computing $ \hat{y}^{(i)} = \hat{f}_{\mathrm{PR}} ( \xi^{(i)} ) $.
More information on polynomial regression can be found in \cite{HaTiFr09} and \cite{Al14}.

As an example, Fig.~\ref{fig: poly rbf} shows the approximation of the test function $ f(x) = \sin( \frac{2\pi}{100}x ) $ (solid red) by a polynomial of degree $ L = 3 $ (blue dots).
To compute the polynomial coefficients, $ D = 100 $ data points $ \{ ( i, \sin(i) ) \}_{i=1}^{D} $ were used, i.e., in this case, we have $ \xi^{(i)} = i $ and $ y^{(i)} = f( \xi^{(i)} ) = \sin( \frac{2\pi}{100}i) $.
While the approximation seems to be acceptable for $ x \in [ 1, 100 ] $, it is poor for $ x > 100 $.
The author of \cite{Mag98} addresses further problems of polynomial regression.

\subsection{Ridge Regression}
In statistics, ridge regression is a synonym for Tikhonov regularization \cite{Ha98}.
Often, this method is applied if solving a system of equations like $ \mb{X} \mb{\theta} = \mb{y} $ (cf.~\eqref{eq: pol reg system of eq}) is not well posed.
In the general case, a regularization term is introduced to form the optimization problem
	\begin{equation}
	\label{eq: rid reg OP}
	\hat{\mb{\theta}}_{\mathrm{RR}} = \arg \min_{\mb{\theta}} \| \mb{y} - \mb{X} \mb{\theta} \|_2^2 + \lambda^2 \| \mb{L} \mb{\theta} \|_2^2
	\end{equation}
where $ \lambda $ is the regularization parameter (cf.~\cite{Ha98}).
The matrix $ \mb{L} $ allows to represent prior assumptions on the size of the solution by the norm $ \| \mb{L} \mb{\theta} \|_2 $.
In most cases, $ \mb{L} = \id $ is a good choice.

An alternative formulation of the problem is to minimize
$ ( \mb{X}^\tp \mb{X} + \lambda^2 \mb{L}^\tp \mb{L} ) \mb{\theta} = \mb{X}^\tp \mb{y} $ (cf.~\cite{Ha98}).
If the intersection of the null-spaces of $ \mb{X} $ and $ \mb{L} $ contains $ \zero $ only, the unique solution is given by
	\begin{equation}
	\label{eq: rid reg solution}
	\hat{\mb{\theta}}_{\mathrm{RR}} = ( \mb{X}^\tp \mb{X} + \lambda^2 \mb{L}^\tp \mb{L} )^{-1} \mb{X}^\tp \mb{y}
	\end{equation}
which is derived in \cite{Ha98}.

For our purposes, ridge regression fits the weights $ \theta_k $ of the function
$ f_{\mathrm{RR}}(x) = \sum_{k=0}^{K} \theta_k g_k(x) $
by means of the minimization in \eqref{eq: rid reg OP}.
The $ g_k(x) $ denote arbitrary functions.
So, e.g., they can be sinusoidal or exponential functions.
The structure of $ \mb{X} $ therefore looks as follows:
	\begin{equation}
	\label{eq: rid reg structure of X}
	\mb{X} =
	\begin{bmatrix}
	g_0(\xi^{(1)}) & g_1(\xi^{(1)}) & \dots & g_K(\xi^{(1)}) \\
	g_0(\xi^{(2)}) & g_1(\xi^{(2)}) & \dots & g_K(\xi^{(2)}) \\
	\vdots & \vdots & & \vdots \\
	g_0(\xi^{(D)}) & g_1(\xi^{(D)}) & \dots & g_K(\xi^{(D)}) 
	\end{bmatrix}.
	\end{equation}
Note that for $ \lambda = 0 $, $ K = L $, and $ g_k(x) = x^k $ equation \eqref{eq: rid reg solution} corresponds to \eqref{eq: pol reg solution}, the least squares polynomial regression.

We predict data with
	\begin{equation}
	\label{eq: rid reg equation}
	\hat{y}^{(i)} = \hat{f}_{\mathrm{RR}}( \xi^{(i)} ) = \sum_{k=0}^{K} \hat{\theta}_{k,\mathrm{RR}} g_k( \xi^{(i)} ). 
	\end{equation}
More information on ridge regression can be found in \cite{Ha98} and \cite{Ma70}.

As the functions $ g_k $ can be chosen arbitrarily, ridge regression allows to incorporate some knowledge of the input signal into the approximation.
Considering that, the test function $ f(x) = \sin(\frac{2\pi}{100}x) $ is an interesting task.
If we know the input function can be modeled by a sinusoidal signal, we can first compute the frequency from the given data and then use an according sine function as $ g_0( x ) $ in \eqref{eq: rid reg structure of X}.
In this special case, the choice $ K = 1, $ $ \lambda = 0 $ leads to no deviation between the approximation and $ f(x) $.

\subsection{Radial Basis Function Network}
According to \cite{HoStWh89}, multilayer feedforward networks with only one hidden layer can approximate any measurable function with arbitrary accuracy.
This makes feedforward networks an interesting tool for approximation.

The radial basis function (RBF) network is a single hidden layer feedforward network (cf.~\cite{KaTi99}).
The output transfer functions are linear, the hidden-layer transfer functions $ g_n(z) $ are nonlinear.
Typically, we have Gaussian $ g_n(z) =  \exp( - \frac{z}{2 \sigma^2} ) $.
The network output is then given by
	\begin{equation}
	\label{eq: rbf equation}
	f_{\mathrm{RB}}( \mb{x} ) = b_0 + \sum_{n=1}^{N} \theta_{n,\mathrm{RB}} g_n( \| \mb{x} - \mb{c}_i \|_2^2 )
	\end{equation}
where $ b_0 $ is a bias term, the $ \mb{c}_i $ are called centers, and $ N $ denotes the number of basis functions (hidden neurons) (cf.~\cite{KaTi99}).
The bias is sometimes incorporated into the summation by introducing an extra constant basis function $ g_0 (z) = 1 $.

There are different training schemes for RBF networks.
A simple way is to fix the $ \mb{c}_i $ in advance.
For example, this can be done by evenly spacing the centers over the input space or choosing the centers equal to the input vectors.
If the scaling parameters $ \sigma $ are also fixed, only the weights $ \theta_{n,\mathrm{RB}} $ have to be determined.
This can be done by minimizing the squared error between the network output $ f_{\mathrm{RB}} (\mb{\xi}^{(i)}) $ and the desired $ y^{(i)} $ (cf.~\cite{ScKePa01}).
More information on RBF networks can be found in \cite{KaTi99,ScKePa01,BrLo88}.

\begin{figure}
	\centering
	\input{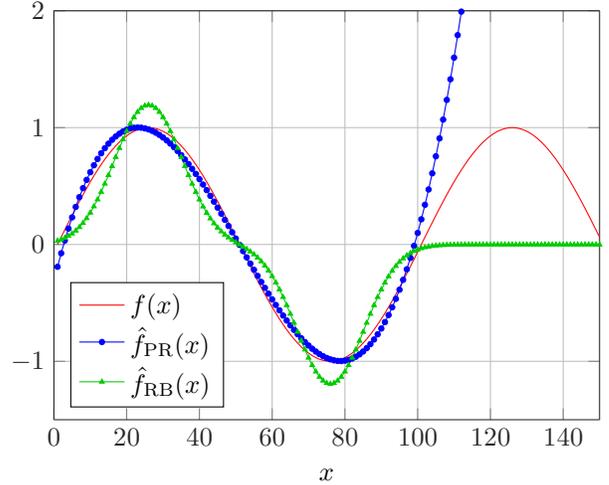}
	\caption{Test signal $ f(x) = \sin(\frac{2\pi}{100}x) $ (solid red) with approximations polynomial regression (PR, blue dots) and radial basis function network (RB, green triangles). $ D = 100 $ training samples were used.}
	\label{fig: poly rbf}
\end{figure}

Fig.~\ref{fig: poly rbf} illustrates the principle of RBF networks (green triangles).
We chose $ N = 2 $ Gaussian basis functions with $ \sigma = 10 $, which can be recognized in the graph.
By increasing $ N $, the approximation can be improved.
As argued in \cite{PaSa91}, RBF networks are capable of universal approximation. 

\subsection{Smoothing Splines}
The spline basis approach aims at finding a twice continuously differentiable function $ \hat{f}_{\mathrm{SS}}(x) $ that minimizes
	\begin{equation}
	\label{eq: smo spl OP}
	\sum_{d=1}^D ( y^{(i)} - f_{\mathrm{SS}}( \xi^{(i)} ) )^2 + \lambda \int( f_{\mathrm{SS}}^{\prime\prime}(t) )^2 \mathrm{d}t
	\end{equation}
(cf.~\cite{HaTiFr09}).
Usually, the smoothing parameter $ \lambda $ is positive.
With $ \lambda = 0 $, the minimizer can be any interpolating function.

The unique minimizer of \eqref{eq: smo spl OP} is known to be a natural cubic spline (cf.~\cite{HaTiFr09}).
A natural cubic spline $ f_{\mathrm{SS}}(x) $ with $ D $ knots can be represented by $ D $ basis functions $ N_d(x) $ as
	\begin{equation}
	\label{eq: smo spl function}
	f_{\mathrm{SS}}(x) = \sum_{d=1}^{D} \theta_d N_d(x).
	\end{equation}
The basis functions are given by
$ N_1(x) = 1 $, $ N_2(x) = x $, $ N_{d+2}(x) = \Delta_d(x) - \Delta_{D-1}(x) $
with
	\begin{equation}
	\Delta_d(x) = \frac{(x-\xi^{(d)})_+^3 - (x-\xi^{(D)})_+^3 }{\xi^{(D)} - \xi^{(d)}}
	\end{equation}
where the data points $ \xi^{(d)} $ are knots and $ (.)_+ $ denotes the positive part (cf.~\cite{HaTiFr09}).

To find the optimal weights in \eqref{eq: smo spl function}, we define the matrices $ [\mb{X}]_{ij} = N_j( \xi^{(i)} ) $ and $ [\mb{\Omega}]_{jk} = \int N_j^{\prime\prime}(t) N_k^{\prime\prime}(t) \mathrm{d}t $ for $ i,j,k = 1, 2, \dots, D $.
This allows us to formulate the optimization problem
	\begin{equation}
	\hat{\mb{\theta}}_{\mathrm{SS}} = \arg \min_{\mb{\theta}} ( \mb{y} - \mb{X} \mb{\theta} )^\tp ( \mb{y} - \mb{X} \mb{\theta} ) + \lambda \mb{\theta}^\tp \mb{\Omega} \mb{\theta}.
	\end{equation}
The solution is then given by $ \hat{\mb{\theta}}_{\mathrm{SS}} = ( \mb{X}^\tp \mb{X} + \lambda \mb{\Omega} )^{-1} \mb{X}^\tp \mb{y} $ (cf.~\cite{HaTiFr09}).
As for large $ D $ the matrices get big, the authors of \cite{HaTiFr09} suggest to apply thinning strategies to simplify the computations.
That is, not all internal knots are used.
The effect on the fit is negligible.

Having found $ \hat{\mb{\theta}}_{\mathrm{SS}} $, we predict data with
	\begin{equation}
	\hat{y}^{(i)} = \hat{f}_{\mathrm{SS}}( \xi^{(i)} ) = \sum_{d=1}^{D} \hat{\theta}_{d,\mathrm{SS}} N_d( \xi^{(i)} ).
	\end{equation}
	
Fig.~\ref{fig: kernel splines tree} shows the result (green triangles) of this method applied to the test function $ f( x ) = \sin( \frac{2\pi}{100}x ) $.
Increasing the smoothing parameter $ \lambda $ puts more emphasis on the second derivative penalty in \eqref{eq: smo spl OP}, i.e., smooths the fit.

\subsection{Kernel Regression}
Kernel regression is also known as kernel smoothing.
The basic idea was introduced by Nadaraya \cite{Na64} and Watson \cite{Wa64}.
The approximating function can be written as
	\begin{equation}
	\hat{f}_{\mathrm{KR}}( \mb{x} ) = \frac	{\sum\limits_{i=1}^{D} y^{(i)} \left( \prod\limits_{j=1}^{P} K_{\lambda}( x_j, \xi_j^{(i)} ) \right)}
											{\sum\limits_{j=1}^{D} \left( \prod\limits_{j=1}^{P} K_{\lambda}( x_j, \xi_j^{(i)} ) \right)}
	\end{equation}
where $ K_\lambda (x,\xi_i) $ is a kernel parameterized by the smoothing parameter $ \lambda $ (cf.~\cite{ShIbKhJaPa10}).
Each kernel needs to meet three properties:
\begin{align}
	\text{1)} \, &K_{\lambda}(x,\xi_i) \geq 0\\
	\text{2)} \, &\int xK_{\lambda}(0,x)\mathrm{d}x = 0\\
	\text{3)} \, &0 < \int x^2 K_{\lambda}(0,x) \mathrm{d}x < \infty.
\end{align}

A common choice is the Gaussian kernel
	\begin{equation}
	K_\lambda (x,\xi_i) = \frac{1}{\sqrt{2\pi}} \exp\left(\frac{-(x-\xi_i)^2}{2\lambda^2}\right).
	\end{equation}
More kernel functions can be found in \cite{ShIbKhJaPa10}.
However, choosing an appropriate kernel is not as important as properly selecting the bandwidth $ \lambda > 0 $.
With increasing $ \lambda $ the approximation smooths.
Setting the bandwidth equal to the variance of all $ y^{(i)} $ is a simple approach (see~\cite{ShIbKhJaPa10}).

Typically, $ \hat{f}_{\mathrm{KR}}( x ) $ shows bad behavior on the boundaries.
This can also be observed in Fig.~\ref{fig: kernel splines tree} (blue dots) for $ x $ near 0 and 100.
To overcome this issue, local polynomial fits at the borders have been suggested (see, e.g., \cite{HaTiFr09}).
For the signal in Fig.~\ref{fig: kernel splines tree}, we selected a Gaussian kernel and bandwidth $ \lambda = 4 $.

\subsection{Autoregressive Integrated Moving Average}
The autoregressive integrated moving average (ARIMA) model is a linear non-stationary model.
In this context, non-stationarity means that the data does not move around a fixed mean.
The idea of an integrated model is to reduce this non-stationary behavior by a differentiation step.
After this, an autoregressive moving average (ARMA) model is applied to the differentiated data (cf.~\cite{BoJeRe12}).

ARIMA is often used in time series analysis.
For this reason, we treat our data as a time series $ y[t] $ with time index $ t $. 
We define two operators to simplify the notation (similar to~\cite{BoJeRe12}).
The backward shift or lag operator $ B $ shifts the time index: $ B^m y[t] = y[t-m] $.
And the backward difference operator $ \nabla $ enables a short notation for the difference between two time series values: $ \nabla y[t] = y[t] - y[t-1] = (1-B) y[t] $.
Before we discuss the general ARIMA model, we first introduce its two components, namely autoregressive and moving average processes, separately.

The value of an autoregressive process of order $ p $ at time $ t $ depends on a linear combination of $ p $ previous values \cite{BoJeRe12}:
	\begin{equation}
	\tilde{y}[t] = \phi_1 \tilde{y}[t-1] + \phi_2 \tilde{y}[t-2] + \dots + \phi_p \tilde{y}[t-p] + a[t]
	\end{equation}
where $ a[t] $ is a random shock from a white noise process with variance $ \sigma_a^2 $.
For convenience, the series $ \tilde{y}[t] = y[t] - \mu $ of deviation from the mean $ \mu $ is introduced.
Using the autoregressive operator $ \phi_p (B) = 1 - \phi_1 B - \phi_2 B^2 - \dots - \phi_p B^p $ of order $ p $, the model can be written as $ \phi_p (B) \tilde{y}[t] = a[t] $ (cf.~\cite{BoJeRe12}).
This model has $ p+2 $ unknown parameters: $ \mu $, $ \sigma_a^2 $, and the $ \phi_i $.

In contrast, the value of a moving average process of order $ q $ at time $ t $ depends on a linear combination of $ q $ previous random shocks \cite{BoJeRe12}:
	\begin{equation}
	\tilde{y}[t] = a[t] - \theta_1 a[t-1] - \theta_2 a[t-2] - \dots - \theta_q a[t-q].
	\end{equation}
Using the moving average operator $ \theta_q (B) = 1 - \theta_1 B - \theta_2 B^2 - \dots - \theta_q B^q $, the model is given by $ \tilde{y}[t] = \theta_q (B) a[t] $.
It has $ q+2 $ unknown parameters: $ \mu $, $ \sigma_a^2 $, and the $ \theta_i $.

The general ARIMA($ p, d, q $) model is formulated as
	\begin{equation}
	\label{eq: ARIMA model}
	\varphi(B) y[t] =  \phi_p (B) \nabla^d y[t] = \theta_0 + \theta_q (B) a[t]
	\end{equation}
where we use the non-stationarity operator $ \varphi(B) = \phi_p (B) \nabla^d = 1 - \varphi_1 B - \varphi_2 B^2 - \dots - \varphi_{p+d} B^{p+d} $ (cf.~\cite{BoJeRe12}).
With $ \theta_0 \neq 0 $ a deterministic polynomial trend of order $ d $ can be realized.
We can also express the model by means of the difference equation:
	\begin{multline}
	\label{eq: ARIMA difference equation}
	y[t] = \varphi_1 y[t-1] + \dots + \varphi_{p+d} y[t-p-d] \\- \theta_1 a[t-1] - \dots - \theta_q a[t-q] + a[t].
	\end{multline}

To determine a forecast $ \hat{y}[t+l] $ for lead time $ l > 0 $ at origin $ t $, the mean squared error $ \ep[ ( y[t+l] - \hat{y}[t+l] )^2 ] $ is minimized.
If the $ a[t] $ form a sequence of independent random variables, it can be shown that the MMSE forecast is given by the conditional expectation
\begin{equation}
	\hat{y}[t+l] = \ep[ y[t+l] | y[t], y[t-1], \dots ] = \ep_t[y[t+l]]
\end{equation}
where $ \ep_t[y[t+l]] $ is a short notation for the conditional expectation.
We can then use the difference equation \eqref{eq: ARIMA difference equation} to compute a forecast:
	\begin{multline}
	\label{eq: ARIMA function}
	\ep_t[ y[t+l] ] = \varphi_1 \ep_t[ y[t+l-1] ] + \dots + \varphi_{p+d} \ep_t[ y[t+l-p-d] ] \\- \theta_1 \ep_t[ a[t+l-1] ] - \dots - \theta_q \ep_t[ a[t+l-q] ] + \ep_t[ a[t+l] ].
	\end{multline}
The conditional expectations in \eqref{eq: ARIMA function} are computed by inserting $ y $'s and $ a $'s when they are given.
Future $ a $'s are set to zero, and intermediate values $ \ep_t[ y[t+1] ] $, $ \ep_t[ y[t+2] ] $, ..., $ \ep_t[ y[t+l-1] ] $ are calculated iteratively.
Using the difference equation \eqref{eq: ARIMA difference equation} is only one way to obtain forecasts.
Two more possibilities are given in \cite{BoJeRe12} which may be advantageous in certain situations.

A general method for obtaining initial parameter estimates of an ARIMA process is described in \cite{BoJeRe12}.
In a preparation step, a new series $ w[t] = \nabla^d y[t] $ of differences is generated, then, a three stage procedure is applied.
However, an outline of the whole method is beyond the scope of this report as much context and many definitions are needed.
The interested reader is referred to the chapters on model identification in \cite{BoJeRe12}.

In some applications, a seasonal behavior of the data can be observed.
For example, if we model hourly changes of the temperature, we expect similarities between measurements that are 24 hours apart.
To incorporate such a periodicity in an ARIMA model, the authors of \cite{BoJeRe12} introduce the so-called multiplicative model
	\begin{equation}
	\phi_p (B) \Phi_P (B^s) \nabla^d \nabla_s^D y[t] = \theta_q (B) \Theta_Q (B^s) a_t
	\end{equation}
as an extension of \eqref{eq: ARIMA model}.
Here, $ s $ describes the periodicity.
So, e.g., in the above temperature example we have $ s = 24 $.
The operator $ \nabla_s^D $ is defined by $ \nabla_s = 1 - B^s $, and $ \Phi_P (B^s) $ and $ \Theta_Q (B^s) $ are polynomials in $ B^s $ of orders $ P $ and $ Q $, respectively.
Forecasting can again be done by means of the according difference equation.
For more information, please refer to the chapter on seasonal models in \cite{BoJeRe12}.

In the example with the test function $ f(x) = \sin(\frac{2\pi}{100}x) $,
it turns out to be sufficient to use an ARIMA(2,0,0) model to predict one period of the sine with a mean squared error of less than $ 10^{-10} $.

\subsection{Regression Tree}
The following overview is based on the considerations in \cite{HaTiFr09}.
The idea of binary regression trees is to partition the input space into $ M $ rectangular regions $ \mathcal{R}_m $ and apply a simple fit, e.g., a constant, to each region.
Using the indicator function
	\begin{equation}
	\mathbb{I}_{\mathcal{R}_m} ( \mb{x} ) =
	\begin{cases}
	1 \quad \text{if } \mb{x} \in \mathcal{R}_m \\
	0 \quad \text{otherwise}
	\end{cases}
	\end{equation}
the output of a regression tree is given by
	\begin{equation}
	f_{\mathrm{RT}}( \mb{x} ) = \sum_{m=1}^{M} c_m \mathbb{I}_{\mathcal{R}_m}(\mb{x})
	\end{equation}
i.e., the constant $ c_m $ of the region $ \mathcal{R}_m $ to which the element $ \mb{x} $ belongs.
By minimizing the sum of squares  $ \sum_{i=1}^{D} ( y^{(i)} - f_{\mathrm{RT}}( \mb{\xi}^{(i)} ) )^2 $, the constant $ c_m $ associated with region $ \mathcal{R}_m $ is determined as the average of all $ y^{(i)} $ in that region: $ c_m = \mathrm{average}( y^{(i)} | \mb{\xi}^{(i)} \in \mathcal{R}_m ) $ (cf.~\cite{HaTiFr09}).
Thus, computing the constants is simple, and the main challenge lies in determining the regions.

We describe a greedy algorithm from \cite{HaTiFr09} to achieve binary partitions in the following.
Let $ x_j $ be a splitting variable and $ s $ a split joint.
We define the two half spaces $ \mathcal{R}_1(j,s) = \{ \mb{x} | x_j \leq s \} $ and $ \mathcal{R}_2(j,s) = \{ \mb{x} | x_j > s \} $.
Then, $ j $ and $ s $ are found by solving
	\begin{equation}
	\min_{j,s} ( \min_{c_1} \hspace{-1mm} \sum_{i: \mb{\xi}^{(i)} \in \mathcal{R}_1(j,s)} \hspace{-2mm} (y^{(i)} - c_1)^2 + \min_{c_2} \hspace{-1mm} \sum_{i: \mb{\xi}^{(i)} \in \mathcal{R}_2(j,s)} \hspace{-2mm} (y^{(i)} - c_2)^2 ).
	\end{equation}
For any $ j $ and $ s $, the solution to the inner problem is given by
$ \hat{c}_1 = \mathrm{average} ( y^{(i)} | \mb{\xi}^{(i)} \in R_1(j,s) ) $ and $ \hat{c}_2 = \mathrm{average} ( y^{(i)} | \mb{\xi}^{(i)} \in R_2(j,s) ) $.
Hence, computing the split joint can be done quickly if $ j $ is given.
By browsing through all inputs, the best pair of $ j $ and $ s $ is found.
The result of this partitioning are two regions on which the same process is repeated.

Basically, this is done until a predetermined tree size is reached.
However, it might be difficult to choose a proper tree size.
One strategy is to first compute a tree $ T_0 $ with a minimum node-size and then prune it by means of cost-complexity pruning  which we illustrate briefly.
By pruning $ T_0 $, i.e., collapsing internal nodes, we get a subtree $ T \subset T_0 $.
Let $ | T | $ be the number of terminal nodes (regions $ \mathcal{R}_m $) of $ T $.
Further, denote by $ N_m $ the number of elements $ \mb{\xi}^{(i)} $ in region $ R_m $.
Additionally, we define the following two quantities:
	\begin{align}
	\hat{c}_m &:= \frac{1}{N_m} \sum_{i: \mb{\xi}^{(i)} \in \mathcal{R}_m} y^{(i)}\\
	Q_m(T) &:= \frac{1}{N_m} \sum_{i: \mb{\xi}^{(i)} \in \mathcal{R}_m} ( y^{(i)} - \hat{c}_m )^2.
	\end{align}
Finally, our cost complexity criterion is defined as
	\begin{equation}
	C_{\alpha}(T) = \sum_{m=1}^{|T|} N_m Q_m(T) + \alpha |T|.
	\end{equation}
For each $ \alpha $, we aim at finding the subtree $ T_\alpha \subseteq T_0 $ which minimizes the cost criterion.
The resulting subtree is unique.
With the parameter $ \alpha \geq 0 $, we control the tree size and fit quality.
Large values lead to small trees, whereas $ \alpha = 0 $ is the original tree $ T_0 $.
The parameter $ \alpha $ can be chosen in accordance to preferences or it can be estimated by cross-validation (cf.~\cite{HaTiFr09}).

One technique to reduce the variance of the prediction is called bagging (cf.~\cite{HaTiFr09})
Basically, the number of given training data is artificially increased by random sampling with replacement.
Given a set $ \mathcal{S} $ of $ D $ training samples, we generate $ B $ sets $ \mathcal{S}_1, \mathcal{S}_2, \dots, \mathcal{S}_B $ with $ D $ data points each.
Then, a tree model $ f_{\mathrm{RT}}^{(b)}( x ) $ is fit to every set $ \mathcal{S}_b $ and the bagging estimate $ f_{\mathrm{bag}}( x ) $ is defined by
	\begin{equation}
	\label{eq: reg tree bagging}
	f_{\mathrm{bag}}( x ) = \frac{1}{B} \sum_{b=1}^{B} f_{\mathrm{RT}}^{(b)}( x ).
	\end{equation}

\begin{figure}
	\centering
	\input{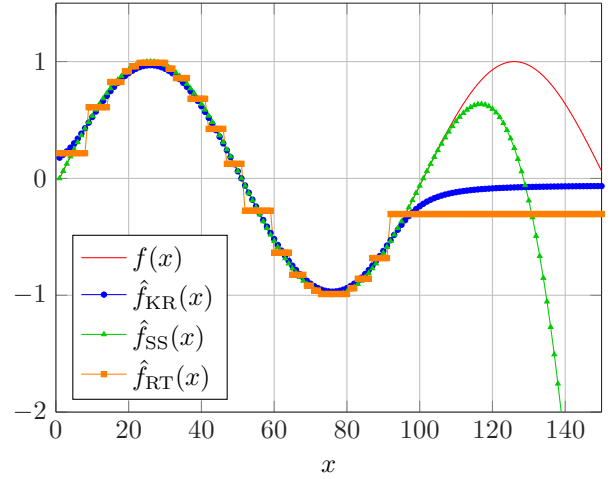}
	\caption{Test signal $ f(x) = \sin(\frac{2\pi}{100}x) $ (solid red) with approximations kernel regression (KR, blue dots), smoothing splines (SS, green triangles), and regression tree (RT, orange squares). $ D = 100 $ training samples were used.}
	\label{fig: kernel splines tree}
\end{figure}

In Fig.~\ref{fig: kernel splines tree}, the regression tree approximation (orange squares) of $ f(x) = \sin(\frac{2\pi}{100}x) $ is given, where $ D = 100 $ training samples were used.
Note that the regions $ \mathcal{R}_m $ and the constant fits can be recognized.
As can be seen, evaluating the tree for $ x > 100 $ gives the constant of the rightmost region.
Thus, in the given one-dimensional case a regression tree without further modifications is not a good choice to forecast data.

\subsection{Nexting}
Basically, Nexting, as introduced in \cite{MoWhSu11}, treats the given data as a time series.
Assume, we have $ P $ signals $ y^{(i)}[t] $, $ i = 1, 2, \dots, P $, where $ t $ denotes the time.
The idea of Nexting is to approximate the return
	\begin{equation}
	\label{eq: nex return}
	G^{(i)}[t] := \sum_{k=0}^{\infty} (\gamma^{(i)})^k y^{(i)}[t+k+1]
	\end{equation}
of signal $ i $ by a function $ f_{\mathrm{nexting}}^{(i)}[t] $.
The discount-rate $ \gamma^{(i)} $ controls the timescale.
For $ \gamma^{(i)} = 0 $, the return $ G^{(i)}[t] = y^{(i)}[t+1] $ reduces to the value in the immediate next time step, which is very short-sighted.
With $ \gamma^{(i)} > 0 $ all (discounted) future signal values are taken into account.

The approximating function has the form
	\begin{equation}
	\label{eq: nex function}
	\hat{f}_{\mathrm{nexting}}^{(i)}(t) = \mb{\phi}^\tp[t] \mb{\theta}^{(i)}[t].
	\end{equation}
This is the inner product of the feature vector $ \mb{\phi}[t] $ and the weight vector $ \mb{\theta}^{(i)}[t] $.
Note that there is only one feature vector for all $ P $ signals.
This vector is constructed by tile coding after normalizing all signals to lie in the interval $ [ 0, 1 ] $ (cf.~\cite{MoWhSu11}).
As a result, $ \mb{\phi}[t] \in \{ 0, 1 \}^N $ has a constant number of 1 features.

The weight vectors $ \mb{\theta}^{(i)}[t] $ are determined by a linear temporal-difference approach \cite{MoWhSu11}:
	\begin{multline}
	\mb{\theta}^{(i)}[t+1] = \mb{\theta}^{(i)}[t] + \alpha (y^{(i)}[t+1] \\ + \gamma^{(i)} \mb{\phi}^\tp[t+1] \mb{\theta}^{(i)}[t] - \mb{\phi}^\tp[t] \mb{\theta}^{(i)}[t]) \mb{e}^{(i)}[t].
	\end{multline}
The so-called step-size parameter $ \alpha $ is positive and the eligibility vector $ \mb{e}^{(i)}[t] $ follows the update rule $ \mb{e}^{(i)}[t] = \gamma^{(i)} \lambda \mb{e}^{(i)}[t-1] + \mb{\phi}[t] $.
Here, $ \lambda \in [0,1] $ is the trace-decay parameter.
A special case is $ \lambda = 1 $.
For this choice, $ \mb{\theta}^{(i)}[t] $ converges asymptotically to the vector which minimizes the mean squared error between the prediction and $ G^{(i)}[t] $.

\begin{figure}
	\centering
	\input{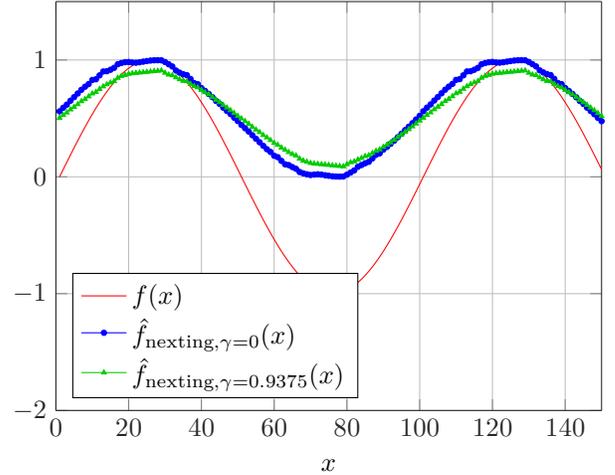}
	\caption{Test signal $ f(x) = \sin(\frac{2\pi}{100}x) $ (solid red) with Nexting approximations. Multiple periods were used for training.}
	\label{fig: nexting}
\end{figure}

For the choice $ \alpha = 0.1 $, $ \lambda = 0.9 $ and the two $ \gamma $-values $ \gamma = 0 $ and $ \gamma = 0.9375 $, Fig.~\ref{fig: nexting} depicts the result of Nexting for the sinusoid signal $ f(x) = \sin(\frac{2\pi}{100}x) $.
As stated above, choosing $ \gamma = 0 $ should lead to a prediction of the function value in the immediate next time step.
That is, $ \hat{f}_{\mathrm{nexting},\gamma = 0}( x ) $ is a forecast of $ f( x+1 ) $.
However, in the situation shown in Fig.~\ref{fig: nexting}, this behavior is hardly recognizable.
Similarly, larger values of $ \gamma $ should indicate changes in $ f( x ) $ earlier.
Again, this can hardly be recognized in our example.
We remark, however, that in Section~\ref{sec: application}, where we apply Nexting to weather data, the expected forecasting behavior can be observed.

Note that the principle of Nexting differs from the algorithms we introduced so far.
According to \eqref{eq: nex function}, in order to compute $ \hat{f}_{\mathrm{nexting}}( x ) $ for $ x > 100 $, we use $ \mb{\phi}[ x ] $ and $ \mb{\theta}[ x ] $, i.e., the prediction $ \hat{f}_{\mathrm{nexting}}( x ) $ is calculated online.
In contrast, polynomial regression, for example, gives a function $ \hat{f}_{\mathrm{PR}}( x ) $ which we evaluate for $ x > 100 $ (cf.~\eqref{eq: pol reg function}).
Thus, Nexting forecasts are based on more input data than, e.g., those of polynomial regression which makes comparisons difficult.
Further, in the example with the sinusoid input signal of Fig.~\ref{fig: nexting}, Nexting requires several training periods to reach a steady state.
For the plot, we used 10 training periods and thus a set of $ D = 1000 $ data points.

\section{Applying the Algorithms to Weather Data}
\label{sec: application}
In the previous section, we had $ f(x) = \sin(\frac{2\pi}{100}x) $ as an example function to illustrate the idea of the algorithms we use.
In this section, we take hourly measurements of wind speed, dry-bulb temperature, and direct normal irradiance as input.
The data points stem from a typical meteorological year 3 (TMY3) set provided by the national renewable energy laboratory \cite{WeatherData} where we chose the weather station of Los Angeles.

In a first step, we apply the methods from Section~\ref{sec: algorithms} to the three sets of weather measurements in order to compare the different algorithms.
Following that, we take a closer look at Nexting and regression trees.

In Fig.~\ref{fig: kernel splines tree} (blue dots), we already saw the critical behavior of kernel smoothing at the boundaries.
This method is thus not suited for data forecasting and therefore excluded from the following considerations.

\subsection{Comparison}
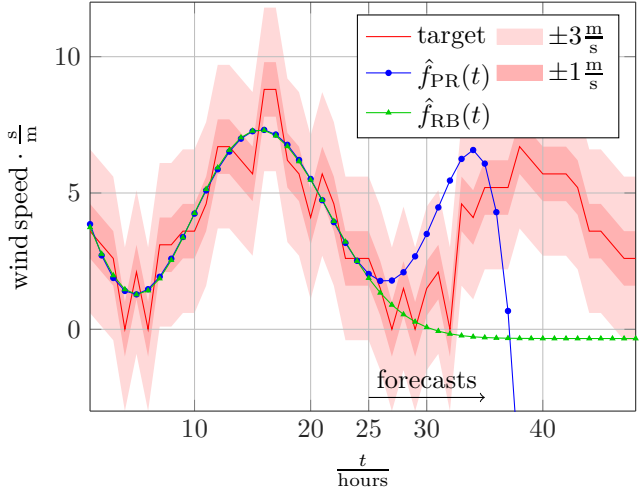
\begin{figure}
	\centering
%
%
%
\definecolor{mycolor1}{rgb}{0.00000,0.44700,0.74100}%
\begin{tikzpicture}
[fill between/on layer={axis background}]
\begin{axis}[%
width=0.4\textwidth,
height=0.3\textwidth,
scale only axis,
separate axis lines,
every outer x axis line/.append style={white!15!black},
every x tick label/.append style={font=\color{white!15!black}},
xmin=1,
xmax=48,
xmajorgrids,
xlabel near ticks,
ylabel near ticks,
xlabel={$ \frac{t}{\mathrm{hours}}$},
every outer y axis line/.append style={white!15!black},
every y tick label/.append style={font=\color{white!15!black}},
ymin=-3,
ymax=12,
ymajorgrids,
ylabel={$ \mathrm{wind\,speed} \cdot \frac{\mathrm{s}}{\mathrm{m}} $},
extra x ticks={25},
extra x tick style={xticklabel=\pgfmathprintnumber{\tick}},
legend style={draw=black,fill=white,legend cell align=left,legend pos=north east},
legend columns=3,
transpose legend,
]
\addplot [color=black!3!white,solid,forget plot,draw=none,name path=C]
  table[row sep=crcr]{%
1	6.6\\
2	6.1\\
3	5.6\\
4	3\\
5	5.1\\
6	3\\
7	6.1\\
8	6.1\\
9	6.6\\
10	6.6\\
11	7.6\\
12	9.7\\
13	9.7\\
14	9.2\\
15	8.7\\
16	11.8\\
17	11.8\\
18	9.2\\
19	8.7\\
20	7.1\\
21	8.7\\
22	7.6\\
23	5.6\\
24	5.6\\
25	5.6\\
26	4.5\\
27	3\\
28	4.5\\
29	3\\
30	4.5\\
31	5.1\\
32	3\\
33	7.6\\
34	7.1\\
35	8.2\\
36	8.2\\
37	8.2\\
38	9.7\\
39	9.2\\
40	8.7\\
41	8.7\\
42	8.7\\
43	8.2\\
44	6.6\\
45	6.6\\
46	6.1\\
47	5.6\\
48	5.6\\
};
\addplot [color=black!3!white,solid,forget plot,draw=none,name path=D]
  table[row sep=crcr]{%
1	0.6\\
2	0.1\\
3	-0.4\\
4	-3\\
5	-0.9\\
6	-3\\
7	0.1\\
8	0.1\\
9	0.6\\
10	0.6\\
11	1.6\\
12	3.7\\
13	3.7\\
14	3.2\\
15	2.7\\
16	5.8\\
17	5.8\\
18	3.2\\
19	2.7\\
20	1.1\\
21	2.7\\
22	1.6\\
23	-0.4\\
24	-0.4\\
25	-0.4\\
26	-1.5\\
27	-3\\
28	-1.5\\
29	-3\\
30	-1.5\\
31	-0.9\\
32	-3\\
33	1.6\\
34	1.1\\
35	2.2\\
36	2.2\\
37	2.2\\
38	3.7\\
39	3.2\\
40	2.7\\
41	2.7\\
42	2.7\\
43	2.2\\
44	0.6\\
45	0.6\\
46	0.1\\
47	-0.4\\
48	-0.4\\
};
\addplot [color=black!10!white,solid,forget plot,draw=none,name path=A]
  table[row sep=crcr]{%
1	4.6\\
2	4.1\\
3	3.6\\
4	1\\
5	3.1\\
6	1\\
7	4.1\\
8	4.1\\
9	4.6\\
10	4.6\\
11	5.6\\
12	7.7\\
13	7.7\\
14	7.2\\
15	6.7\\
16	9.8\\
17	9.8\\
18	7.2\\
19	6.7\\
20	5.1\\
21	6.7\\
22	5.6\\
23	3.6\\
24	3.6\\
25	3.6\\
26	2.5\\
27	1\\
28	2.5\\
29	1\\
30	2.5\\
31	3.1\\
32	1\\
33	5.6\\
34	5.1\\
35	6.2\\
36	6.2\\
37	6.2\\
38	7.7\\
39	7.2\\
40	6.7\\
41	6.7\\
42	6.7\\
43	6.2\\
44	4.6\\
45	4.6\\
46	4.1\\
47	3.6\\
48	3.6\\
};
\addplot [color=black!10!white,solid,forget plot,draw=none,name path=B]
  table[row sep=crcr]{%
1	2.6\\
2	2.1\\
3	1.6\\
4	-1\\
5	1.1\\
6	-1\\
7	2.1\\
8	2.1\\
9	2.6\\
10	2.6\\
11	3.6\\
12	5.7\\
13	5.7\\
14	5.2\\
15	4.7\\
16	7.8\\
17	7.8\\
18	5.2\\
19	4.7\\
20	3.1\\
21	4.7\\
22	3.6\\
23	1.6\\
24	1.6\\
25	1.6\\
26	0.5\\
27	-1\\
28	0.5\\
29	-1\\
30	0.5\\
31	1.1\\
32	-1\\
33	3.6\\
34	3.1\\
35	4.2\\
36	4.2\\
37	4.2\\
38	5.7\\
39	5.2\\
40	4.7\\
41	4.7\\
42	4.7\\
43	4.2\\
44	2.6\\
45	2.6\\
46	2.1\\
47	1.6\\
48	1.6\\
};
\addplot [color=red,solid]	
  table[row sep=crcr]{%
1	3.6\\
2	3.1\\
3	2.6\\
4	0\\
5	2.1\\
6	0\\
7	3.1\\
8	3.1\\
9	3.6\\
10	3.6\\
11	4.6\\
12	6.7\\
13	6.7\\
14	6.2\\
15	5.7\\
16	8.8\\
17	8.8\\
18	6.2\\
19	5.7\\
20	4.1\\
21	5.7\\
22	4.6\\
23	2.6\\
24	2.6\\
25	2.6\\
26	1.5\\
27	0\\
28	1.5\\
29	0\\
30	1.5\\
31	2.1\\
32	0\\
33	4.6\\
34	4.1\\
35	5.2\\
36	5.2\\
37	5.2\\
38	6.7\\
39	6.2\\
40	5.7\\
41	5.7\\
42	5.7\\
43	5.2\\
44	3.6\\
45	3.6\\
46	3.1\\
47	2.6\\
48	2.6\\
};
\addlegendentry{target};
\addplot [color=blue,solid,mark=*,mark size=1]	
  table[row sep=crcr]{%
1	3.85660742705564\\
2	2.70918283166111\\
3	1.88299775636661\\
4	1.40816022408379\\
5	1.28077333710768\\
6	1.47054708254999\\
7	1.92777056282309\\
8	2.58964465117457\\
9	3.38597507227246\\
10	4.24422590784107\\
11	5.09393352734748\\
12	5.87048094373867\\
13	6.51823259422922\\
14	6.99302954613973\\
15	7.26404512778577\\
16	7.31500098441763\\
17	7.14474355921045\\
18	6.76718099930526\\
19	6.21058048690043\\
20	5.51622599539388\\
21	4.73643647057591\\
22	3.93194443687249\\
23	3.16863502863958\\
24	2.5136454465076\\
25	2.03082483877674\\
26	1.77555460786344\\
27	1.78892914179606\\
28	2.09129697076289\\
29	2.67516234870978\\
30	3.49744725998871\\
31	4.4711138510564\\
32	5.45614728722501\\
33	6.24989903446163\\
34	6.57679056623977\\
35	6.07737749544092\\
36	4.29677413130651\\
37	0.67243846144234\\
38	-5.47868244113045\\
39	-14.9736465858688\\
40	-28.7776508075552\\
41	-48.018804084089\\
42	-74.0035404292337\\
43	-108.232671360309\\
44	-152.418077940831\\
45	-208.50004239811\\
46	-278.665219315787\\
47	-365.365246401323\\
48	-471.335994828446\\
};
\addlegendentry{$ \hat{f}_{\mathrm{PR}}( t ) $};
\addplot [color=green!80!black,solid,mark=triangle*,mark size=1]	
table[row sep=crcr]{%
	1	3.74145639359777\\
	2	2.76402838519343\\
	3	1.97580128140027\\
	4	1.46434867987236\\
	5	1.27772350843955\\
	6	1.42123966994426\\
	7	1.86149107082336\\
	8	2.53566914068224\\
	9	3.36340581772133\\
	10	4.25843511703972\\
	11	5.13814671793057\\
	12	5.93020377186399\\
	13	6.57638551413983\\
	14	7.034394578422\\
	15	7.27845704511675\\
	16	7.29926183023898\\
	17	7.10337056618988\\
	18	6.71191216537889\\
	19	6.15829520850187\\
	20	5.48483288664493\\
	21	4.73847899365913\\
	22	3.96617278573068\\
	23	3.21045929151603\\
	24	2.50602957995223\\
	25	1.87762931532628\\
	26	1.33948783174186\\
	27	0.896119352961772\\
	28	0.544123670863684\\
	29	0.274508972340893\\
	30	0.0750755306420583\\
	31	-0.0674959400868789\\
	32	-0.16605999322703\\
	33	-0.231989595612325\\
	34	-0.274677710158216\\
	35	-0.301442148217159\\
	36	-0.317696545706956\\
	37	-0.327261068328534\\
	38	-0.332715354044203\\
	39	-0.335730331680886\\
	40	-0.337346105000974\\
	41	-0.338185750775509\\
	42	-0.338608898883723\\
	43	-0.338815733781048\\
	44	-0.338913804413915\\
	45	-0.338958915310628\\
	46	-0.338979047472935\\
	47	-0.338987765158126\\
	48	-0.338991428232039\\
};
\addlegendentry{$ \hat{f}_{\mathrm{RB}}( t ) $};
\addplot[red!15!white] fill between[of=C and D];
\addlegendentry{$ \pm 3 \frac{\mathrm{m}}{\mathrm{s}} $};
\addplot[red!30!white] fill between[of=A and B];
\addlegendentry{$ \pm 1 \frac{\mathrm{m}}{\mathrm{s}} $};
\draw [->] (axis cs:25,-2.5) -- node[above] {forecasts} (axis cs:35,-2.5);
\end{axis}
\end{tikzpicture}%
	\caption{Wind speed target signal (solid red) approximated by polynomial regression (PR, blue dots) and radial basis function network (RB, green triangles). $ D = 24 $ samples were used for training. The shaded areas represent a deviation of $ \pm 1 \frac{\mathrm{m}}{\mathrm{s}} $ and $ \pm 3 \frac{\mathrm{m}}{\mathrm{s}} $ around the target signal.}
	\label{fig: wind example}
\end{figure}

For a qualitative statement of the forecast accuracy, we define intervals of a constant deviation around the input target signals.
As an illustration, the shaded areas in Fig.~\ref{fig: wind example} represent a deviation of $ \pm 1 \frac{\mathrm{m}}{\mathrm{s}} $ and $ \pm 3 \frac{\mathrm{m}}{\mathrm{s}} $ around the wind speed.
We try to train all algorithms such that as many predictions as possible lie within the smaller interval.
The idea is to gain an understanding of the algorithms' performance.

We use $ D = 24 $ data samples (one day) to train the algorithms and then forecast one day.
Tab.~\ref{tab: results all} lists our findings.
The first column of the three data sets represents the root mean square error (RMSE) between the actual values and the approximations for the 24 training samples.
Column two then gives the number of consecutive forecast samples (the number of hours) which lie within the two intervals defined above.
The settings which lead to the table entries can be found in Tab.~\ref{tab: settings all}.

As an example, Fig.~\ref{fig: wind example} depicts the polynomial regression and radial basis function (RBF) network approximations of the wind speed (solid red).
From \eqref{eq: rbf equation} we know that RBF networks approach a constant bias term for large $ x $-values.
In the depicted case, the wind speed curve shape coincides with the decaying behavior of the approximation which is the reason for the seemingly good forecast. 
In general, RBF networks are more fit to interpolate than to forecast data.

\begin{figure}
	\centering
%
%
%
\definecolor{mycolor1}{rgb}{0.00000,0.44700,0.74100}%
\begin{tikzpicture}
[fill between/on layer={axis background}]
\begin{axis}[%
width=0.4\textwidth,
height=0.3\textwidth,
scale only axis,
separate axis lines,
every outer x axis line/.append style={white!15!black},
every x tick label/.append style={font=\color{white!15!black}},
xmin=1,
xmax=48,
xmajorgrids,
xlabel near ticks,
ylabel near ticks,
xlabel={$ \frac{t}{\mathrm{hours}}$},
every outer y axis line/.append style={white!15!black},
every y tick label/.append style={font=\color{white!15!black}},
ymin=12.5,
ymax=23,
ymajorgrids,
ylabel={$ \mathrm{temperature} \cdot \frac{1}{^\circ\mathrm{C}} $},
extra x ticks={25},
extra x tick style={xticklabel=\pgfmathprintnumber{\tick}},
legend style={draw=black,fill=white,legend cell align=left,legend pos=north west},
legend columns=3,
transpose legend,
]
\addplot [color=black!3!white,solid,forget plot,draw=none,name path=C]
  table[row sep=crcr]{%
1	17.1\\
2	17.1\\
3	17.1\\
4	17.6\\
5	17.6\\
6	18.2\\
7	17.6\\
8	18.2\\
9	18.7\\
10	19.8\\
11	20.9\\
12	20.9\\
13	20.9\\
14	20.4\\
15	19.3\\
16	19.8\\
17	19.3\\
18	17.6\\
19	17.1\\
20	16.5\\
21	15.9\\
22	15.9\\
23	15.9\\
24	16.5\\
25	16.5\\
26	16.5\\
27	16.5\\
28	16.5\\
29	17.1\\
30	17.6\\
31	17.6\\
32	18.7\\
33	18.7\\
34	18.7\\
35	20.9\\
36	20.9\\
37	21.5\\
38	20.9\\
39	20.4\\
40	20.4\\
41	19.8\\
42	18.7\\
43	18.2\\
44	17.6\\
45	17.1\\
46	17.1\\
47	17.1\\
48	17.1\\
};
\addplot [color=black!3!white,solid,forget plot,draw=none,name path=D]
  table[row sep=crcr]{%
1	14.1\\
2	14.1\\
3	14.1\\
4	14.6\\
5	14.6\\
6	15.2\\
7	14.6\\
8	15.2\\
9	15.7\\
10	16.8\\
11	17.9\\
12	17.9\\
13	17.9\\
14	17.4\\
15	16.3\\
16	16.8\\
17	16.3\\
18	14.6\\
19	14.1\\
20	13.5\\
21	12.9\\
22	12.9\\
23	12.9\\
24	13.5\\
25	13.5\\
26	13.5\\
27	13.5\\
28	13.5\\
29	14.1\\
30	14.6\\
31	14.6\\
32	15.7\\
33	15.7\\
34	15.7\\
35	17.9\\
36	17.9\\
37	18.5\\
38	17.9\\
39	17.4\\
40	17.4\\
41	16.8\\
42	15.7\\
43	15.2\\
44	14.6\\
45	14.1\\
46	14.1\\
47	14.1\\
48	14.1\\
};
\addplot [color=black!10!white,solid,forget plot,draw=none,name path=A]
  table[row sep=crcr]{%
1	16.1\\
2	16.1\\
3	16.1\\
4	16.6\\
5	16.6\\
6	17.2\\
7	16.6\\
8	17.2\\
9	17.7\\
10	18.8\\
11	19.9\\
12	19.9\\
13	19.9\\
14	19.4\\
15	18.3\\
16	18.8\\
17	18.3\\
18	16.6\\
19	16.1\\
20	15.5\\
21	14.9\\
22	14.9\\
23	14.9\\
24	15.5\\
25	15.5\\
26	15.5\\
27	15.5\\
28	15.5\\
29	16.1\\
30	16.6\\
31	16.6\\
32	17.7\\
33	17.7\\
34	17.7\\
35	19.9\\
36	19.9\\
37	20.5\\
38	19.9\\
39	19.4\\
40	19.4\\
41	18.8\\
42	17.7\\
43	17.2\\
44	16.6\\
45	16.1\\
46	16.1\\
47	16.1\\
48	16.1\\
};
\addplot [color=black!10!white,solid,forget plot,draw=none,name path=B]
  table[row sep=crcr]{%
1	15.1\\
2	15.1\\
3	15.1\\
4	15.6\\
5	15.6\\
6	16.2\\
7	15.6\\
8	16.2\\
9	16.7\\
10	17.8\\
11	18.9\\
12	18.9\\
13	18.9\\
14	18.4\\
15	17.3\\
16	17.8\\
17	17.3\\
18	15.6\\
19	15.1\\
20	14.5\\
21	13.9\\
22	13.9\\
23	13.9\\
24	14.5\\
25	14.5\\
26	14.5\\
27	14.5\\
28	14.5\\
29	15.1\\
30	15.6\\
31	15.6\\
32	16.7\\
33	16.7\\
34	16.7\\
35	18.9\\
36	18.9\\
37	19.5\\
38	18.9\\
39	18.4\\
40	18.4\\
41	17.8\\
42	16.7\\
43	16.2\\
44	15.6\\
45	15.1\\
46	15.1\\
47	15.1\\
48	15.1\\
};
\addplot [color=red,solid]	
  table[row sep=crcr]{%
1	15.6\\
2	15.6\\
3	15.6\\
4	16.1\\
5	16.1\\
6	16.7\\
7	16.1\\
8	16.7\\
9	17.2\\
10	18.3\\
11	19.4\\
12	19.4\\
13	19.4\\
14	18.9\\
15	17.8\\
16	18.3\\
17	17.8\\
18	16.1\\
19	15.6\\
20	15\\
21	14.4\\
22	14.4\\
23	14.4\\
24	15\\
25	15\\
26	15\\
27	15\\
28	15\\
29	15.6\\
30	16.1\\
31	16.1\\
32	17.2\\
33	17.2\\
34	17.2\\
35	19.4\\
36	19.4\\
37	20\\
38	19.4\\
39	18.9\\
40	18.9\\
41	18.3\\
42	17.2\\
43	16.7\\
44	16.1\\
45	15.6\\
46	15.6\\
47	15.6\\
48	15.6\\
};
\addlegendentry{target};
\addplot [color=blue,solid,mark=*,mark size=1] 	
  table[row sep=crcr]{%
1	15.5916540925365\\
2	15.5816680609696\\
3	15.6856943357555\\
4	15.9971342571909\\
5	16.240089730268\\
6	16.4481275020004\\
7	16.3121131856294\\
8	16.6150575212317\\
9	17.287241639301\\
10	18.3263051539019\\
11	19.2460056252031\\
12	19.4959874798536\\
13	19.3487032535561\\
14	18.7788408183906\\
15	18.1450191923228\\
16	18.1240478609159\\
17	17.5877938954037\\
18	16.3597834813697\\
19	15.538531264563\\
20	14.9467867483236\\
21	14.4728072894157\\
22	14.3407076658151\\
23	14.4767300358281\\
24	14.953169910254\\
25	15.42960978468\\
26	15.4377487616462\\
27	14.5092859436928\\
28	12.1759204333602\\
29	7.9693513331885\\
30	1.421277745718\\
31	-7.93660122651109\\
32	-20.5725864809585\\
33	-36.954978915084\\
34	-57.5520794263474\\
};
\addlegendentry{$ \hat{f}_{\mathrm{SS}}( t ) $};
\addplot [color=green!80!black,solid,mark=triangle*,mark size=1]	
table[row sep=crcr]{%
	1	14.6445184922653\\
	2	14.7132795447053\\
	3	14.9148767499345\\
	4	15.2355715915717\\
	5	15.6535092461326\\
	6	16.1402079531336\\
	7	16.6625\\
	8	17.1847920468664\\
	9	17.6714907538674\\
	10	18.0894284084283\\
	11	18.4101232500655\\
	12	18.6117204552947\\
	13	18.6804815077347\\
	14	18.6117204552947\\
	15	18.4101232500655\\
	16	18.0894284084283\\
	17	17.6714907538674\\
	18	17.1847920468664\\
	19	16.6625\\
	20	16.1402079531336\\
	21	15.6535092461326\\
	22	15.2355715915717\\
	23	14.9148767499345\\
	24	14.7132795447053\\
	25	14.6445184922653\\
	26	14.7132795447053\\
	27	14.9148767499345\\
	28	15.2355715915717\\
	29	15.6535092461326\\
	30	16.1402079531336\\
	31	16.6625\\
	32	17.1847920468664\\
	33	17.6714907538674\\
	34	18.0894284084283\\
	35	18.4101232500655\\
	36	18.6117204552947\\
	37	18.6804815077347\\
	38	18.6117204552947\\
	39	18.4101232500655\\
	40	18.0894284084283\\
	41	17.6714907538674\\
	42	17.1847920468664\\
	43	16.6625\\
	44	16.1402079531336\\
	45	15.6535092461326\\
	46	15.2355715915717\\
	47	14.9148767499345\\
	48	14.7132795447053\\
};
\addlegendentry{$ \hat{f}_{\mathrm{RR}}( t ) $};
\addplot[red!15!white] fill between[of=C and D];
\addlegendentry{$ \pm 1.5 ^\circ \mathrm{C} $};
\addplot[red!30!white] fill between[of=A and B];
\addlegendentry{$ \pm 0.5 ^\circ \mathrm{C} $};
\draw [->] (axis cs:25,13) -- node[above] {forecasts} (axis cs:35,13);
\end{axis}
\end{tikzpicture}%
	\caption{Dry-bulb temperature target signal (solid red) approximated by smoothing splines (SS, blue dots) and ridge regression (RR, green triangles). $ D = 24 $ samples were used for training. The shaded areas represent a deviation of $ \pm 0.5 ^\circ \mathrm{C} $ and $ \pm 1.5 ^\circ \mathrm{C} $ around the target signal.}
	\label{fig: temp example}
\end{figure}
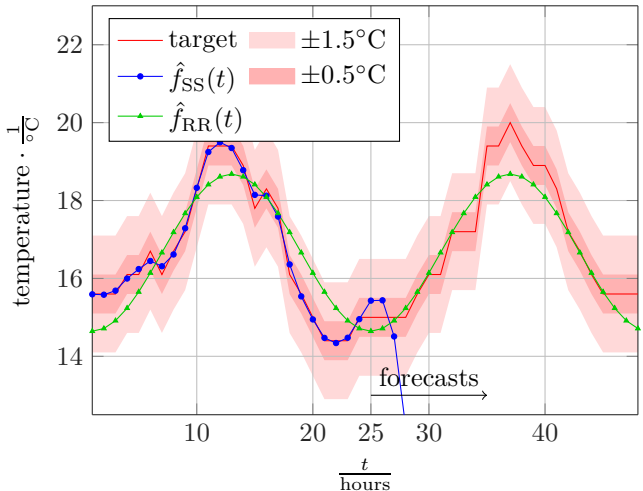

There are many ways to select the functions $ g_k $ in \eqref{eq: rid reg equation} for ridge regression.
We chose a constant $ g_0 $ and sinusoid $ g_1 $ for the three data sets.
Fig.~\ref{fig: temp example} shows the result for this simple choice on the example of the dry-bulb temperature.
While the training RMSE is comparatively large, it is interesting to see that a sinusoidal function indicates the trend fairly well.

As we saw in Fig.~\ref{fig: kernel splines tree}, regression trees basically split up the abscissa into a number of intervals and fit a constant to these.
Forecasting then gives the value of the rightmost interval which is in general not very accurate.
Instead, one idea might be to view the tree as a prototype representation for one signal period.
To make predictions, we then evaluate this prototype period.
So, e.g., to forecast the temperature for $ t = 26 $, we evaluate the prototype at $ t = 2 $, which was done in Tab.~\ref{tab: results all} for all data sets.
In other words, we perform a modulo 24 operation on $ t $ prior to evaluating the tree.

Following the idea of tree bagging (see \eqref{eq: reg tree bagging}), the described prototype approach can be used to incorporate multiple training periods in the predictions.
We investigate the benefit of this in the next subsection.

Tab.~\ref{tab: results all} also presents the results of a seasonal ARIMA model applied to the three data sets wind speed, dry-bulb temperature, and direct normal irradiance.
However, to train a model with seasonality $ s = 24 $, we had to use two days of training data.
Further, it is difficult to determine the orders $ p, q, P, Q $ of the associated polynomials.
The sample autocorrelation and partial autocorrelation functions give a first idea (cf.~\cite{BoJeRe12}).
For our data, we combined the insights delivered by these functions with the practical guideline that orders of $ P, Q > 1 $ are rarely necessary (cf.~\cite{BoJeRe12}) and compared the result of different parameter values.

We already mentioned the difficulty that arises in comparing Nexting to the other algorithms in Sec.~\ref{sec: algorithms}.
In our studies, several training periods were necessary to reach a steady state.
Thus, the bad performance of Nexting in Tab.~\ref{tab: results all} where only one period was used to train is not surprising.
For the results in Tab.~\ref{tab: results all}, we fixed the weight vector $ \mb{\theta} $ in \eqref{eq: nex function} after 24 training samples, i.e., the algorithm's learning process was stopped after 24 training samples.
This makes a comparison to the other algorithms more fair.

\subsection{Nexting and Regression Trees with Multiple Training Periods}
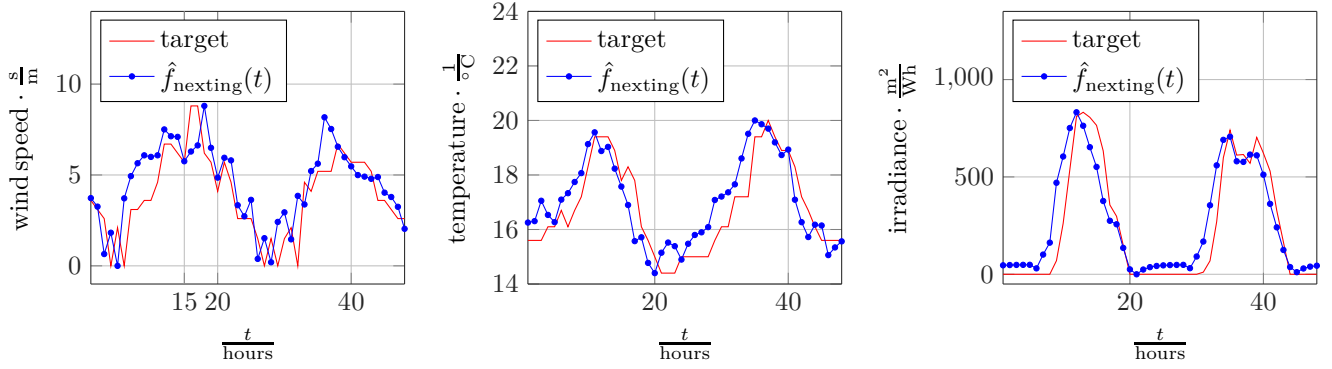
\begin{figure*}
	\centering
%
%
%
\definecolor{mycolor1}{rgb}{0.00000,0.44700,0.74100}%
\definecolor{mycolor2}{rgb}{0.85000,0.32500,0.09800}%
\begin{tikzpicture}

\begin{axis}[%
width=0.23\textwidth,
height=0.2\textwidth,
scale only axis,
separate axis lines,
every outer x axis line/.append style={white!15!black},
every x tick label/.append style={font=\color{white!15!black}},
xmin=1,
xmax=48,
xmajorgrids,
xlabel near ticks,
ylabel near ticks,
xlabel={$ \frac{t}{\mathrm{hours}}$},
every outer y axis line/.append style={white!15!black},
every y tick label/.append style={font=\color{white!15!black}},
ymin=-1,
ymax=14,
ymajorgrids,
ylabel={$ \mathrm{wind\,speed} \cdot \frac{\mathrm{s}}{\mathrm{m}} $},
extra x ticks={15},
extra x tick style={xticklabel=\pgfmathprintnumber{\tick}},
legend style={draw=black,fill=white,legend cell align=left,legend pos=north west}
]
\addplot [color=red,solid]
  table[row sep=crcr]{%
1	3.6\\
2	3.1\\
3	2.6\\
4	0\\
5	2.1\\
6	0\\
7	3.1\\
8	3.1\\
9	3.6\\
10	3.6\\
11	4.6\\
12	6.7\\
13	6.7\\
14	6.2\\
15	5.7\\
16	8.8\\
17	8.8\\
18	6.2\\
19	5.7\\
20	4.1\\
21	5.7\\
22	4.6\\
23	2.6\\
24	2.6\\
25	2.6\\
26	1.5\\
27	0\\
28	1.5\\
29	0\\
30	1.5\\
31	2.1\\
32	0\\
33	4.6\\
34	4.1\\
35	5.2\\
36	5.2\\
37	5.2\\
38	6.7\\
39	6.2\\
40	5.7\\
41	5.7\\
42	5.7\\
43	5.2\\
44	3.6\\
45	3.6\\
46	3.1\\
47	2.6\\
48	2.6\\
};
\addlegendentry{target};
\addplot [color=blue,mark=*,mark size=1,solid]	
  table[row sep=crcr]{%
1	3.7274522878271\\
2	3.25861805055802\\
3	0.650166996062629\\
4	1.82203150882973\\
5	0\\
6	3.71410736097046\\
7	4.93632532477408\\
8	5.6454898858806\\
9	6.08062826398287\\
10	5.99458813147814\\
11	6.08168264465944\\
12	7.50629499101996\\
13	7.13085547365717\\
14	7.09706164340771\\
15	5.75446102079766\\
16	6.29072164942249\\
17	6.62696720433475\\
18	8.8\\
19	6.49433145051054\\
20	4.85376189202704\\
21	5.94665339599787\\
22	5.80254737605905\\
23	3.33565566719757\\
24	2.7236625838986\\
25	3.63080356722551\\
26	0.383115531046075\\
27	1.52185969288482\\
28	0.197236410592303\\
29	2.41669019594635\\
30	2.94406515697177\\
31	1.4579518448319\\
32	3.85031970551133\\
33	3.37758188603247\\
34	5.21625388884215\\
35	5.62245054933104\\
36	8.17346522422899\\
37	7.52978657564713\\
38	6.55613128674286\\
39	5.97641936252544\\
40	5.47353302872949\\
41	4.99337319529209\\
42	4.90889554549983\\
43	4.78870757012353\\
44	4.88714545051644\\
45	4.02716701307688\\
46	3.78818297402647\\
47	3.24742226509989\\
48	2.0379348989126\\
};
\addlegendentry{$ \hat{f}_{\mathrm{nexting}}( t ) $};
\end{axis}
\end{tikzpicture}%
\quad
%
%
%
\definecolor{mycolor1}{rgb}{0.00000,0.44700,0.74100}%
\begin{tikzpicture}

\begin{axis}[%
width=0.23\textwidth,
height=0.2\textwidth,
scale only axis,
separate axis lines,
every outer x axis line/.append style={white!15!black},
every x tick label/.append style={font=\color{white!15!black}},
xmin=1,
xmax=48,
xmajorgrids,
xlabel near ticks,
ylabel near ticks,
xlabel={$ \frac{t}{\mathrm{hours}}$},
every outer y axis line/.append style={white!15!black},
every y tick label/.append style={font=\color{white!15!black}},
ymin=14,
ymax=24,
ymajorgrids,
ylabel={$ \mathrm{temperature} \cdot \frac{1}{^\circ\mathrm{C}} $},
legend style={draw=black,fill=white,legend cell align=left,legend pos=north west}
]
\addplot [color=red,solid]
  table[row sep=crcr]{%
1	15.6\\
2	15.6\\
3	15.6\\
4	16.1\\
5	16.1\\
6	16.7\\
7	16.1\\
8	16.7\\
9	17.2\\
10	18.3\\
11	19.4\\
12	19.4\\
13	19.4\\
14	18.9\\
15	17.8\\
16	18.3\\
17	17.8\\
18	16.1\\
19	15.6\\
20	15\\
21	14.4\\
22	14.4\\
23	14.4\\
24	15\\
25	15\\
26	15\\
27	15\\
28	15\\
29	15.6\\
30	16.1\\
31	16.1\\
32	17.2\\
33	17.2\\
34	17.2\\
35	19.4\\
36	19.4\\
37	20\\
38	19.4\\
39	18.9\\
40	18.9\\
41	18.3\\
42	17.2\\
43	16.7\\
44	16.1\\
45	15.6\\
46	15.6\\
47	15.6\\
48	15.6\\
};
\addlegendentry{target};
\addplot [color=blue,mark=*,mark size=1,solid]	
  table[row sep=crcr]{%
1	16.2543506331718\\
2	16.306387869427\\
3	17.0541306271625\\
4	16.5345823114427\\
5	16.2697264943249\\
6	17.0962321373573\\
7	17.3320447673143\\
8	17.7421388482492\\
9	18.0685184187008\\
10	19.1326927285512\\
11	19.5637164377859\\
12	18.8787081544013\\
13	19.0301310767977\\
14	18.2325188193324\\
15	17.5745177774737\\
16	16.8983385143319\\
17	15.5733484825911\\
18	15.7138294778082\\
19	14.7742424355673\\
20	14.4\\
21	15.14875816505\\
22	15.521203407114\\
23	15.3875119308513\\
24	14.8972532174421\\
25	15.4766860790023\\
26	15.7994477769307\\
27	15.8964630858405\\
28	16.0872710123603\\
29	17.0830872353398\\
30	17.209034058526\\
31	17.3639353896523\\
32	17.6558878493355\\
33	18.6070987341262\\
34	19.5107365731298\\
35	20\\
36	19.8607857879639\\
37	19.7009446491456\\
38	19.1954487348973\\
39	18.7313174523729\\
40	18.9313447305983\\
41	17.0925452866246\\
42	16.2678562580713\\
43	15.7236914811957\\
44	16.1762608975619\\
45	16.1463976621606\\
46	15.0611074722022\\
47	15.3421225103397\\
48	15.5645274513319\\
};
\addlegendentry{$ \hat{f}_{\mathrm{nexting}}( t ) $};
\end{axis}
\end{tikzpicture}%
\quad
%
%
%
\definecolor{mycolor1}{rgb}{0.00000,0.44700,0.74100}%
\begin{tikzpicture}

\begin{axis}[%
width=0.23\textwidth,
height=0.2\textwidth,
scale only axis,
separate axis lines,
every outer x axis line/.append style={white!15!black},
every x tick label/.append style={font=\color{white!15!black}},
xmin=1,
xmax=48,
xmajorgrids,
xlabel near ticks,
ylabel near ticks,
xlabel={$ \frac{t}{\mathrm{hours}}$},
every outer y axis line/.append style={white!15!black},
every y tick label/.append style={font=\color{white!15!black}},
ymin=-50,
ymax=1350,
ymajorgrids,
ylabel={$ \mathrm{irradiance} \cdot \frac{\mathrm{m}^2}{\mathrm{Wh}} $},
legend style={draw=black,fill=white,legend cell align=left,legend pos=north west}
]
\addplot [color=red,solid]
  table[row sep=crcr]{%
1	0\\
2	0\\
3	0\\
4	0\\
5	0\\
6	0\\
7	0\\
8	0\\
9	70\\
10	261\\
11	537\\
12	810\\
13	832\\
14	806\\
15	765\\
16	634\\
17	356\\
18	298\\
19	149\\
20	0\\
21	0\\
22	0\\
23	0\\
24	0\\
25	0\\
26	0\\
27	0\\
28	0\\
29	0\\
30	0\\
31	11\\
32	69\\
33	270\\
34	599\\
35	740\\
36	612\\
37	615\\
38	570\\
39	703\\
40	622\\
41	530\\
42	327\\
43	165\\
44	0\\
45	0\\
46	0\\
47	0\\
48	0\\
};
\addlegendentry{target};
\addplot [color=blue,mark=*,mark size=1,solid]
  table[row sep=crcr]{%
1	46.0789853565384\\
2	47.4489203066524\\
3	48.1338877817094\\
4	48.4763715192379\\
5	48.6476133880021\\
6	30.425089351653\\
7	101.206134933895\\
8	162.501341499822\\
9	470.052469416133\\
10	604.737041635673\\
11	751.173646289649\\
12	832\\
13	762.722367361824\\
14	652.858839315973\\
15	551.523588693077\\
16	376.385214534813\\
17	274.793665973844\\
18	256.824290973527\\
19	135.533312251815\\
20	25.188280637469\\
21	0\\
22	24.4094276283832\\
23	36.6141414425748\\
24	42.7164983496706\\
25	45.7676768032184\\
26	47.2932660299924\\
27	48.0560606433794\\
28	48.4374579500729\\
29	31.3053671468128\\
30	91.3444674302022\\
31	168.160197771641\\
32	354.389376279917\\
33	559.342734320386\\
34	690.409434148475\\
35	706.568484893523\\
36	580.301863242465\\
37	577.174192928759\\
38	614.727163437097\\
39	610.751573693895\\
40	511.50570524056\\
41	361.700459677975\\
42	241.000790037594\\
43	125.013053694164\\
44	37.0184979021495\\
45	10.8243685639215\\
46	29.8216119103439\\
47	39.3202335835552\\
48	44.0695444201608\\
};
\addlegendentry{$ \hat{f}_{\mathrm{nexting}}( t ) $};
\end{axis}
\end{tikzpicture}%
	\caption{Target signals (solid red) wind speed, dry-bulb temperature, and direct normal irradiance (from left to right) approximated by Nexting (blue dots) with $ \gamma = 0 $. Multiple periods were used for training.}
	\label{fig: nexting multiple periods}
\end{figure*}

As stated above, Nexting achieves better results if applied to multiple training periods.
Fig.~\ref{fig: nexting multiple periods} shows Nexting approximations for the three cases wind speed, dry-bulb temperature, and direct normal irradiance.
In contrast to the previous experiments, we did not fix the weight vector.
Further, 20 training periods as opposed to only one were provided.
The parameter values were again those of Tab.~\ref{tab: settings all} --- except for direct normal irradiance where  $ \alpha = 0.1 $ was chosen.
As by default the Nexting output lies in between zero and one, we scaled it to match the target signal for a better illustration.

The discount rate is $ \gamma = 0 $ in all cases.
Hence, we expect the value of $ \hat{f}_{\mathrm{nexting}}( t ) $ to be a forecast of the target signal at $ t + 1 $ (see \eqref{eq: nex return}).
In Fig.~\ref{fig: nexting multiple periods}, this behavior can generally be observed.
However, there are some points where the approximation lags behind.
An example is given by the left chart of Fig.~\ref{fig: nexting multiple periods} around $ t = 15 $.
The wind speed shows fast variations which seem to be a challenge for Nexting.

In the previous subsection, we suggested to compute regression trees for multiple signal training periods and average the predictions of those as in \eqref{eq: reg tree bagging}.
In combination with the modulo 24 operation which we described above, this might improve the root mean square error.

The best choice for the wind speed as target turns out to be 6 training periods which improves the RMSE to 1.1893 as apposed to 1.2096 in Tab.~\ref{tab: results all}.
Further, using two training periods in the case of direct normal irradiance results in a RMSE of 120.4159 as apposed to 132.4193.
However, there is no improvement for the temperature as target.

\section{Conclusion}
The results of the last paragraph show that regression trees can benefit from multiple training periods.
However, too many periods can worsen the performance and it is difficult to decide for the right number.
Nevertheless, regression trees are suited to forecast a trend as Tab.~\ref{tab: results all} suggests.

We saw that ridge regression requires knowledge of the target signal model to properly select the functions $ g_i $ in \eqref{eq: rid reg structure of X}.
The possibility to incorporate such knowledge can be an advantage.
However, in contrast to smoothing splines and polynomial regression where basically only one parameter needs to be tuned, it might be difficult to decide for suitable functions.
While the computation time for these three methods is about the same, ridge regression requires more preliminary work.

In our studies, choosing appropriate parameters for the seasonal ARIMA models was difficult.
What is more, the computation time was high in comparison to the algorithms mentioned so far.
Further, in contrast to all methods of Sec.~\ref{sec: algorithms}, the ARIMA model does not provide predictions for the training interval.
Nevertheless, it is worthwhile investigating such models in future research as the forecasts were quite accurate.

We already mentioned that radial basis function networks can approximate functions with arbitrary accuracy.
However, as argued above, we deem those not suited for forecasting tasks.
Additionally, the training phase was time-consuming in our experiments.

Lastly, Nexting's online learning strategy is advantageous in our weather data context as new incoming measurements can easily be incorporated to update the forecasts.
However, the output needs to be scaled in order to provide useful numerical values.
Also, in contrast to the other algorithms, Nexting predictions cannot be calculated for arbitrary points in time without further effort.

\begin{table*}
	\centering
	\caption{Wind speed, dry-bulb temperature, and direct normal irradiance approximated by different methods trained with 24 samples. The training root mean square error is given as well as the number of consecutive forecast samples that lie within a interval around the target signal.}
	\begin{tabular}
		{
			|m{15em}?
			 >{\centering\arraybackslash} m{6em}
			|>{\centering\arraybackslash} m{6em}?
			 >{\centering\arraybackslash} m{6em}
			|>{\centering\arraybackslash} m{6em}?
			 >{\centering\arraybackslash} m{6em}
			|>{\centering\arraybackslash} m{6em}|
		}
		\hline
		Method
		& \makecell{Wind Speed\\RMSE}	& \makecell{Wind Speed\\$ \pm 1, \pm 3 \frac{\mathrm{m}}{\mathrm{s}} $}
		& \makecell{Temperature\\RMSE}	& \makecell{Temperature\\$ \pm 0.5, \pm 1.5 ^\circ\mathrm{C} $}
		& \makecell{Irradiance\\RMSE}	& \makecell{Irradiance\\$ \pm 100, \pm 300 \frac{\mathrm{Wh}}{\mathrm{m}^2} $}\\
		\thickhline
		Polynomial Regression
		& 0.9337		& 2, 7				& 0.3467				& 2, 2							& \enskip72.6259	& 1, 1					\\
		\hline
		Ridge Regression
		& 1.0306		& \enskip2, 24		& 0.7389				& \enskip6, 24					& 173.5266			& \enskip5, 24			\\
		\hline
		Radial Basis Function Network
		& 0.9358		& 5, 8				& 0.4149				& 7, 9							& \enskip44.6131	& 8, 9					\\
		\hline
		Smoothing Splines
		& 0.9286		& 3, 5				& 0.1441				& 3, 3							& \enskip25.8115	& 9, 9					\\
		\hline
		Seasonal ARIMA\textsuperscript{1}
		& -				& \enskip2, 15		& -						& \enskip3, 23					& -					& 8, 9					\\
		\hline
		Regression Tree
		& 1.2096		& 2, 7				& 0.6020				& \enskip0, 17					& 132.4193			& \enskip8, 16			\\
		\hline
		Nexting\textsuperscript{2}
		& 2.6362		& 1, 2				& 1.4651				& 2, 8							& 317.3383			& 5, 5					\\
		\hline
		\multicolumn{7}{l}{\textsuperscript{1}\footnotesize{Two training periods were used.}}\\
		\multicolumn{7}{l}{\textsuperscript{2}\footnotesize{The weight vector was fixed after one period. To compute the RMSE, the signal was rescaled and shifted in time.}}
	\end{tabular}
	\label{tab: results all}
\end{table*}

\begin{table*}
	\centering
	\caption{Settings for the Results in Table~\ref{tab: results all}}
	\begin{tabular}{|l|l|c|c|c|}
		\hline
		Method & Parameters	& Wind Speed & Temperature & Irradiance \\
		\hline
		Polynomial Regression
		&	polynomial degree $ L $ 			& 6		& 7		& 6		\\
		\hline
		Ridge Regression
		&	regularization parameter 			& 0.1 	& 0.1	& 5		\\
		&	function $ g_1 $ 					& $ \cos\left( \frac{2\pi}{24}x - \frac{\pi}{4} \right) $
												& $ \cos\left( \frac{2\pi}{24}x \right) $
												& $ \cos\left( \frac{2\pi}{24}x \right) $	\\
		\hline
		Radial Basis Function
		&	number of basis functions $ N $ 	& 4		& 5		& 4		\\
		& 	scaling parameter $ \sigma $		& 6.3	& 6.6	& 3.7	\\
		\hline
		Smoothing Splines
		&	smoothing parameter $ \lambda $ 	& 9		& 0.1	& 0.5	\\
		\hline
		Seasonal ARIMA
		&	AR orders $ p, P $ 					& 0, 1	& 2, 0	& 0, 1	\\
		& 	MA order $ q, Q $ 					& 3, 0	& 0, 0	& 1, 1	\\
		& 	seasonality $ s $ 					& 24	& 24	& 24	\\
		& 	degrees of differencing $ d, D $ 	& 0, 1	& 2, 1	& 0, 1	\\
		\hline
		Regression Tree
		& 	number of nodes 					& 5 	& 5 	& 5 \\
		\hline
		Nexting
		&	discount rate $ \gamma^{(i)} $ 		& 0		& 0 	& 0 	\\
		& 	step-size parameter $ \alpha $ 		& 0.3	& 0.3 	& 0.2	\\
		& 	trace-decay parameter $ \lambda $ 	& 0.9	& 0.9	& 0.9	\\
		\hline
	\end{tabular}
	\label{tab: settings all}
\end{table*}

\bibliographystyle{IEEEtran}
\bibliography{bibfile_forschungspraxis}
\end{document}